\documentclass[final,5p,times,twocolumn]{elsarticle}

\usepackage{times}
\usepackage{epsfig}
\usepackage{graphicx}
\usepackage{amsmath}
\usepackage{amssymb}
\usepackage{xcolor}
\usepackage{colortbl}
\usepackage{subfigure}
\usepackage{amsmath}
\usepackage{amssymb}
\usepackage{caption}
\usepackage{amsfonts}
\usepackage{array}
\usepackage{graphicx}
\usepackage{mathrsfs}
\usepackage{multirow}
\usepackage{makecell}
\usepackage{siunitx}
\usepackage{stfloats}
\usepackage{algpseudocode}
\usepackage{algorithm}
\usepackage{algorithmicx}
\usepackage{color}
\usepackage{booktabs}
\usepackage{adjustbox}

\usepackage[pagebackref=true,breaklinks=true,colorlinks,bookmarks=false]{hyperref}
\newcommand{\Red}[0]{\rowcolor{red}}

\newcommand{\markchanged}[1]{}

\journal{Neurocomputing}

\begin{document}

\begin{frontmatter}

\title{Focal and Efficient IOU Loss for Accurate Bounding Box Regression}%


\author{Yi-Fan Zhang$^{a,b}$, Weiqiang Ren$^c$, Zhang Zhang$^{a,b,c}$, Zhen Jia$^{a,b}$, Liang Wang$^{a,b}$,
        and~Tieniu Tan$^{a,b}$\\
{$^a${CRIPAC \& NLPR, CASIA, Beijing, China}\\
            }
$^b${{University of Chinese Academy of Sciences, Beijing, China}}\\
$^c${{Horizon Robotics.}}\\}
\begin{abstract}
  In object detection, bounding box regression (BBR) is a crucial step that determines the object localization performance. However, we find that most previous loss functions for BBR have two main drawbacks: (i) Both $\ell_n$-norm and IOU-based loss functions are inefficient to depict the objective of BBR, which leads to slow convergence and inaccurate regression results. (ii) Most of the loss functions ignore the imbalance problem in BBR that the large number of anchor boxes which have small overlaps with the target boxes contribute most to the optimization of BBR. To mitigate the adverse effects caused thereby, we perform thorough studies to exploit the potential of BBR losses in this paper. Firstly, an Efficient Intersection over Union (EIOU) loss is proposed, which explicitly measures the discrepancies of three geometric factors in BBR, i.e., the overlap area, the central point and the side length. After that, we state the Effective Example Mining (EEM) problem and propose a regression version of focal loss to make the regression process focus on high-quality anchor boxes. Finally, the above two parts are combined to obtain a new loss function, namely Focal-EIOU loss. Extensive experiments on both synthetic and real datasets are performed. Notable superiorities on both the convergence speed and the localization accuracy can be achieved over other BBR losses.
\end{abstract}




\end{frontmatter}


\section{Introduction}
Object detection, which includes two sub-tasks: object classification and object localization, is always one of the most fundamental problems in computer vision. Current state-of-the-art object detectors (e.g., Cascade R-CNN~\cite{Cascade}, Mask R-CNN~\cite{maskrcnn}, Dynamic R-CNN~\cite{DynamicRCNN}, and {DETR}~\cite{carion2020end}) rely on a bounding box regression (BBR) module to localize objects. Based on this paradigm, a well-designed loss function is of vital importance for the success of BBR. So far, most of loss functions for BBR fall into two categories:
\begin{figure}[t]
\begin{center}
   \includegraphics[width=0.8\linewidth,scale=0.6]{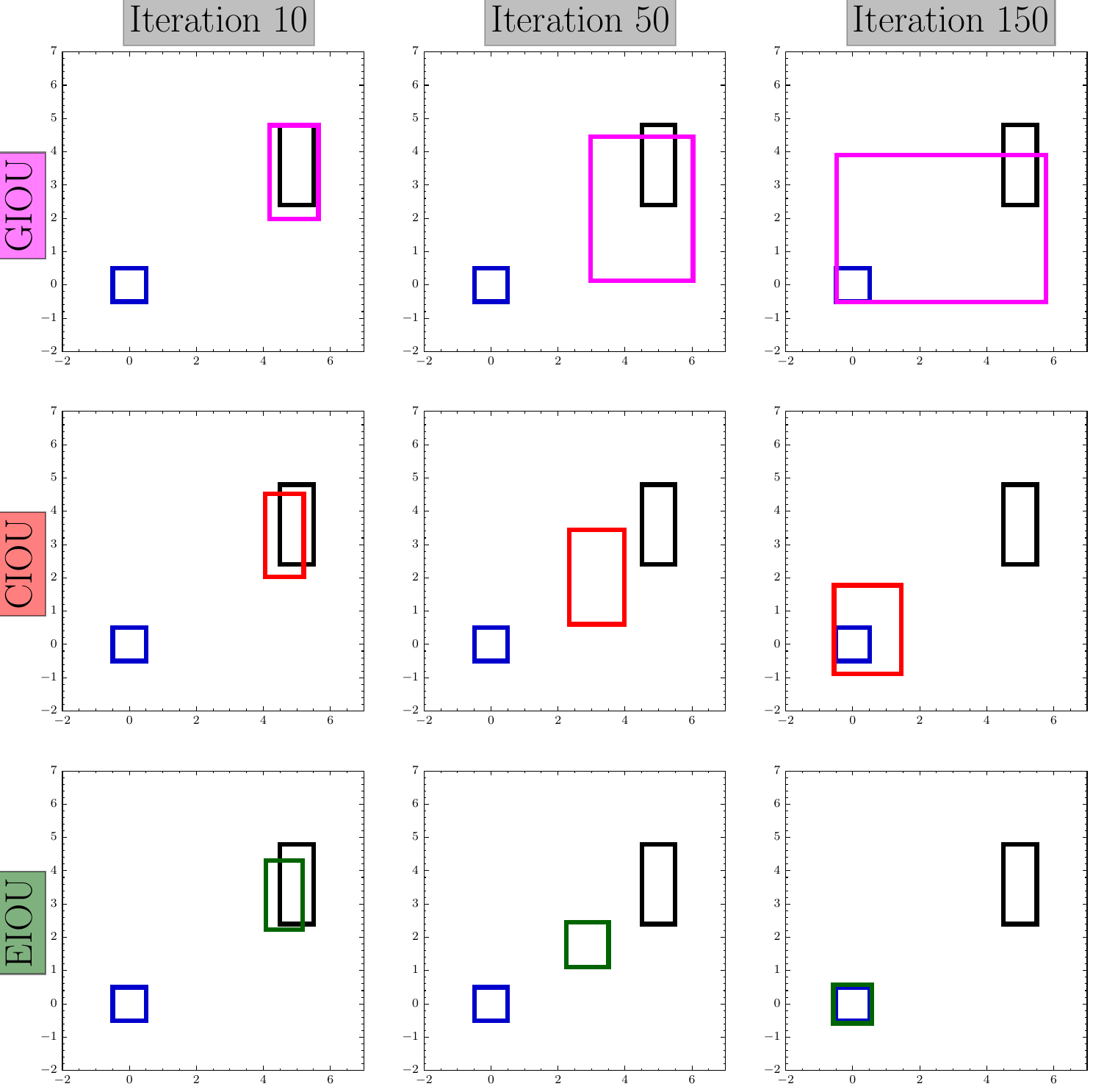}
\end{center}
   \caption{\markchanged{Illustrations on the problems of current BBR losses. Each row shows the optimization results in different iterations with certain loss function. The \textbf{Black} denotes the anchor box. The {\color{blue}Blue} denotes the target box. The fist row denotes {\color{purple}GIOU}. The second row denotes {\color{red}CIOU}. The third row denotes the proposed {\color{green}EIOU}.}}
\label{simple_cmp}
\end{figure}

\begin{itemize}
\item $\ell_n$-norm losses can be unified as Eq.~\eqref{equ:unifiedl1}.
\begin{equation}\label{equ:unifiedl1}
L(x)=\left\{
\begin{array}{ll}
 f(x),  & \text{if }|x|<\beta,\\
 g(x),  & \text{otherwise},\\
\end{array} \right.
\end{equation}
where $x$ is the difference between the predicted box and the target box. Traditional SmoothL1 loss \cite{fast} can be formed as $\beta=1$, $f(x)=0.5|x^2|/\beta$ and $g(x)=|x|-0.5\beta$. $\ell_n$-norm losses have been criticized for not only ignoring the correlations in BBR variables $(x,y,w,h)$ but also the intrinsic bias to the large bounding boxes (due to the unnormalized form)~\cite{2016iou}. However, previous IOU-based losses, e.g., CIOU and GIOU, cannot measure the discrepancies between target box and anchors efficiently, which leads to slow convergences and inaccurate localizations in optimizations of BBR models as illustrated in~\figurename~\ref{simple_cmp}.

\item Intersection over Union (IOU)-based losses can be unified as Eq.~\eqref{equ:unifiedIOU}.
\begin{equation}
L(B, B^{gt}) = 1-\frac{|B\cap B^{gt}|}{|B\cup B^{gt}|}+R(B,B^{gt}),
\label{equ:unifiedIOU}
\end{equation}
where $B$ and $B^{gt}$ are the predicted box and the target box. The penalty term $R(B, B^{gt})$ is designed for the complementary benefit to the original IOU cost. These losses jointly regress all the BBR variables as a whole unit. They are also normalized and insensitive to the scales of bounding boxes. However, most of them suffer from the slow convergence speed and inaccurate localizations. What's more, the existing IOU-based losses neglect the importance of the informative anchor boxes.
\end{itemize}

In this paper, we take thorough studies to exploit the potential of current BBR losses for accurate object detections. Firstly, an Efficient IOU loss (EIOU) is proposed to improve the convergence speed and localization accuracy, which uses an additional penalty term $R(B,B^{gt})$ to explicitly measure the discrepancies of three key geometric factors in BBR, including the overlap area, the central point and the side length. Secondly, we state the Effective Example Mining (EEM) problem in BBR. Inspired by the focal loss \cite{fatal_loss} originally applied to measure the classification errors, we design a regression version of focal loss to enhance the contributions of high-quality anchor boxes with large IOUs in the optimization process of BBR models. Finally, the two proposed methods are combined as a new BBR loss function, namely Focal-EIOU, for efficient and accurate object detection. The effectiveness and advantages of the proposed loss functions are validated by extensive evaluations on both the synthetic and real datasets. Furthermore, when we incorporate the Focal-EIOU loss with several state-of-the-art object detection models, including Faster R-CNN~\cite{faster}, Mask R-CNN ~\cite{maskrcnn}, RetinaNet~\cite{fatal_loss}, ATSS~\cite{ATSS}, PAA~\cite{paa} and {DETR}~\cite{carion2020end}, consistent and significant improvements of detection accuracy can be achieved on the large scale COCO 2017 dataset~\cite{coco}, which illustrates the promising potentials of the proposed loss function.



The contributions of this paper can be summarized as follows:
\begin{enumerate}
  \item Considering the flaws of the IOU-based losses and $\ell_n$-norm losses, we propose an efficient IOU loss to tackle the dilemma of existing losses and obtain a faster convergence speed and superior regression results.
  \item Considering the imbalance between high and low-quality anchor boxes in BBR, we design a regression version of focal loss to enhance contributions of the most promising anchor boxes in model optimization while suppress the irrelevant ones'.
  \item Extensive experiments have been conducted on both synthetic and real data. Outstanding experimental results validate the superiority of the proposed methods. Detailed ablation studies exhibit the effects of different settings of loss functions and parameter values .
\end{enumerate}

\section{Related Work}

\markchanged{In order to improve the accuracy of object detection, some researchers propose methods with cascade structures, such as Cascade R-CNN~\cite{Cascade}, which trains a sequence of detectors with increasing IoU thresholds; RefineDet~\cite{cao2019hierarchical} and IoU-Net~\cite{jiang2018acquisition}, which use multiple fully-convolutional head-networks for predications.  HSD~\cite{cao2019hierarchical} firstly proposes the ROC module, which conducts box regression and classification separately, and a hierarchical shot detector is then proposed by stacking two ROC modules. BorderDet~\cite{qiu2020borderdet} first adopts a simple dense object detector (FCOS) as the baseline to generate the coarse bounding box predictions and then re-extract the features from the second-to-last feature map of FCOS.}
\markchanged{
In addition to network architecture design. There are also some studies who seek more efficient loss functions, which play important roles in the localization results and are orthogonal to model architecture designs. In this section, we briefly survey the related work on loss functions for BBR and the problem of Effective Example Mining (EEM) in object detection.}
\subsection{Loss Functions for BBR}
The regression of bounding boxes is a crucial step in object detection. It aims to refine the location of a predicted bounding box based on the initial proposal or the anchor box.
Till now, BBR has been used on most of the recent detection methods~\cite{2010DPM,rcnn,faster,Wu_2020_CVPR,Beery2020ContextRL,Pang_2019_CVPR,locnet}.

Researchers have spent many efforts in designing loss functions for BBR. YOLO v1~\cite{yolo} proposes to predict the square root of the bounding box size to remedy scale sensitivity. Fast R-CNN~\cite{fast} and Faster R-CNN~\cite{faster} use SmoothL1 loss function, which is a robust $\ell_1$ loss that is less sensitive to outliers than the $\ell_2$ loss used in R-CNN~\cite{rcnn} and SPPNet~\cite{sppnet}. The Dynamic SmoothL1 Loss~\cite{DynamicRCNN} has the same $f(x)$ and $g(x)$ (in Eq.\eqref{equ:unifiedl1}) as the SmoothL1 loss, while it dynamically controls the shape of the loss function to gradually focus on high-quality anchor boxes. The BalanceL1 loss~\cite{Libra} is proposed to redefine $f(x)$ and $g(x)$ to obtain larger gradients for inliers, but the gradients of outliers are not influenced. However, the $\ell_n$-norm loss functions mostly assume the four variables of bounding boxes $(x,y,w,h)$ are independent, which is inconsistent with reality. To address the above problems, the IOU~\cite{2016iou} loss is proposed and it achieves the superior performance on FDDB benchmark at that time. Further, the Generalized IOU (GIOU)~\cite{2019giou} loss is proposed to address the weaknesses of the IOU loss, i.e., the IOU loss will always be zero when two boxes have no interaction. Recently, the Distance IOU and Complete IOU have been proposed~\cite{2020diou}, where the two losses have faster convergence speed and better performance. Pixels IOU ~\cite{2020piou} increases both the angle and IOU accuracy for oriented bounding boxes regression.

{We address the weakness of existing loss functions, then propose an efficient loss function for object detection.}

\subsection{Effective Example Mining}\label{sec:EEM}
One stage detectors suffer from the imbalance issue, such as inter-class imbalance between foreground and background examples. To address this challenge, SSD \cite{ssd} adopts hard negative mining, which only keeps a small set of hard background examples for training. Focal loss~\cite{fatal_loss} re-weights the background and foreground examples such that the hard examples are assigned with large weights. OHEM~\cite{ohem} presents a simple yet surprisingly effective online hard example mining algorithm for training region-based ConvNet detectors. The AP loss~\cite{aploss} and DR loss~\cite{drloss} transform the classification task into the sorting task, to avoid the imbalance between negative and positive examples. In BBR, the imbalance problem still exists, where most anchor boxes have small overlaps with target boxes. While only a small quantity of boxes are most informative for object localization. The most irrelevant boxes with small IOUs will produce excessively large gradients that are inefficient for the training of regression models. Libra R-CNN~\cite{Libra} and Dynamic R-CNN~\cite{DynamicRCNN} suggest that a well-regressed bounding box should contribute more gradients in the model optimization process, based on which they revise the SmoothL1 loss to re-weight predicted bounding boxes.

However, the revised losses~\cite{DynamicRCNN, Libra} can only increase gradients of high-quality examples and cannot suppress the outliers'. Different from the above work, we design a regression version of focal loss to sufficiently exploit the most promising anchor boxes.

\section{Efficient Intersection over Union Loss}
In this section, we firstly analyze the drawbacks of existing popular loss functions and then propose the Efficient IOU loss.

\subsection{Limitations of IOU-Based Losses}\label{subsubsection:drawbackiou}
In this subsection, we analyze the flaws of three IOU-based loss functions, i.e, the IOU~\cite{2016iou}, GIOU~\cite{2019giou} and CIOU~\cite{2020diou} loss.
\subsubsection{Limitations of IOU Loss}
The IOU loss~\cite{2016iou} for measuring similarity between two arbitrary shapes (volumes) $A, B \subseteq \mathbb{S} \in \mathbb{R}^n$ is attained by:
\begin{equation}\label{iou}
  L_{IOU}=1-\frac{|A\cap B|}{|A\cup B|},
\end{equation}
which has some good properties, such as non-negativity, symmetry, triangle inequality and scale insensitivity. It has been proved to be a metric (by the mathematical definition~\cite{iou_analysis}). However, it has two major drawbacks:
\begin{itemize}
  \item If two boxes do not have any intersections, the IOU loss will be always zero, which cannot reflect the closeness between this two boxes correctly.
  \item The convergence speed of the IOU loss is slow.
\end{itemize}
\subsubsection{Limitations of Generalized IOU Loss}
The GIOU loss~\cite{2019giou} loss is proposed to solve the drawbacks of the IOU loss and it is defined as follows,
\begin{equation}\label{giou}
  L_{GIOU}=1-IOU+\frac{|C-(A\cup B)|}{|C|},
\end{equation}
where $A,B\subseteq \mathbb{S}\in \mathbb{R}^n$ are two arbitrary boxes. $C$ is the smallest convex box $C \subseteq \mathbb{S}\in \mathbb{R}^n$ enclosing both $A$ and $B$ and $IOU={|A\cap B|}/{|A\cup B|}$.
The GIOU loss works when $|A\cap B|=0$ while it still has two drawbacks:
\begin{itemize}
  \item When $|A\cap B|=0$, the GIOU loss intends to increase the bounding box's area, making it overlap the target box (see \figurename~\ref{simple_cmp}), which is opposite to the intuition that decreasing the discrepancy of spatial positions.
  \item When $|A\cap B|>0$, the area of $|C-A\cup B|$ is always a small number or equals zero (when $A$ contains $B$, this term will be zero, vice versa). In such a case, the GIOU loss degrades to the IOU loss. As a consequence, the converge rate of the GIOU loss is still slow.
\end{itemize}

\subsubsection{Limitations of Complete IOU Loss}

The CIOU loss~\cite{2020diou} considers three important geometric factors, i.e., the overlap area, the central point distance and the aspect ratio. Given a predicted box $B$ and a target box $B^{gt}$, the CIOU loss is defined as follows.

\begin{equation}\label{CIOU}
  L_{CIOU}=1-IOU+\frac{\rho^2(\mathbf{b},\mathbf{b^{gt}})}{c^2}+\alpha v,
\end{equation}
 where $\mathbf{b}$ and $\mathbf{b}^{gt}$ denote the central points of $B$ and $B^{gt}$ respectively. $\rho(\cdot)=||\mathbf{b}-\mathbf{b}^{gt}||_2$ indicates the Euclidean distance. $c$ is the diagonal length of the smallest enclosing box covering the two boxes. $v=\frac{4}{\pi^2}(\arctan\frac{w^{gt}}{h^{gt}}-\arctan\frac{w}{h})^2$ and $\alpha=\frac{v}{(1-IOU)+v}$ measure the discrepancy of the width-to-height ratio.

The gradient of $v$, w.r.t $w$ and $h$, is calculated as follows.
\begin{equation}\label{CIOU_der}
\begin{aligned}
\frac{\partial v}{\partial w}=\frac{8}{\pi^2}(\arctan\frac{w^{gt}}{h^{gt}}-\arctan\frac{w}{h})*\frac{h}{w^2+h^2},\\
\frac{\partial v}{\partial h}=-\frac{8}{\pi^2}(\arctan\frac{w^{gt}}{h^{gt}}-\arctan\frac{w}{h})*\frac{w}{w^2+h^2}.
\end{aligned}
\end{equation}

In the previous work \cite{2020diou}, experimental results show that both the converge speed and detection accuracy of the CIOU loss have a significant improvement, compared to previous loss functions. However, $v$ in the last term of $L_{CIOU}$ is still not well-defined, which slows down the convergence speed of CIOU from three aspects.

\begin{itemize}
\item In Eq.~\eqref{CIOU}, $v$ just reflects the discrepancy of aspect ratio, rather than the real relations between $w$ and $w^{gt}$ or $h$ and $h^{gt}$. Namely, all the boxes with the property $\{(w=kw^{gt},h=kh^{gt})|k\in R^+\}$ have $v=0$, which is inconsistent with reality.

\item In Eq.~\eqref{CIOU_der}, we have $\frac{\partial v}{\partial w}=-\frac{h}{w}\frac{\partial v}{\partial h}$. $\frac{\partial v}{\partial w}$ and $\frac{\partial v}{\partial h}$ have opposite signs. Thus, at any time, if one of these two variables ($w$ or $h$) is increased, the other one will decrease. It is unreasonable especially when $w<w^{gt}$ and $h<h^{gt}$ or $w>w^{gt}$ and $h>h^{gt}$.

\item {Since the $v$ only reflects the discrepancy of aspect ratio, the CIOU loss may optimizes the similarity in a unreasonable way. As shown in Fig.~\ref{simple_cmp}, the scales of the target box are set as $w^{gt}=1$ and $h^{gt}=1$. The initial scales of the anchor box are set as $w=1$ and $h=2.4$. The anchor box's scales are regressed to $w=1.64$ and $h=2.84$ after $50$ iterations. Here, the CIOU loss indeed increases the similarity of the aspect ratio, while it hinders the model from reducing the true discrepancy between $(w,h)$ and $(w^{gt},h^{gt})$ efficiently.}
\end{itemize}

\subsection{The Proposed Method}
To address the above problems, we revise the CIOU loss and propose a more efficient version of IOU loss, i.e., the EIOU loss, which is defined as follows.
\begin{equation}\label{EIOU}
\begin{aligned}
  L&_{EIOU}=L_{IOU}+ L_{dis} + L_{asp}\\
   &=1-IOU+\frac{\rho^2(\mathbf{b},\mathbf{b^{gt}})}{(w^c)^2+(h^c)^2}+\frac{\rho^2({w},{w^{gt}})}{(w^c)^2}+
  \frac{\rho^2({h},{h^{gt}})}{(h^c)^2},
\end{aligned}
\end{equation}
where $h^w$ and $h^c$ are the width and height of the smallest enclosing box covering the two boxes. Namely, we divide the loss function into three parts: the IOU loss $L_{IOU}$, the distance loss $L_{dis}$ and the aspect loss $L_{asp}$. In this way, we can retain the profitable characteristics of the CIOU loss. At the same time, the EIOU loss directly minimizes the difference of the target box's and anchor box's width and height, which results in a faster converge speed and a better localization result. For a clear demonstration of the superiorities of the EIOU loss, we perform simulation experiments with synthetic data as presented in Section~\ref{sec:sim}.
\begin{figure}[h]
  \begin{center}
    \includegraphics[width=\linewidth]{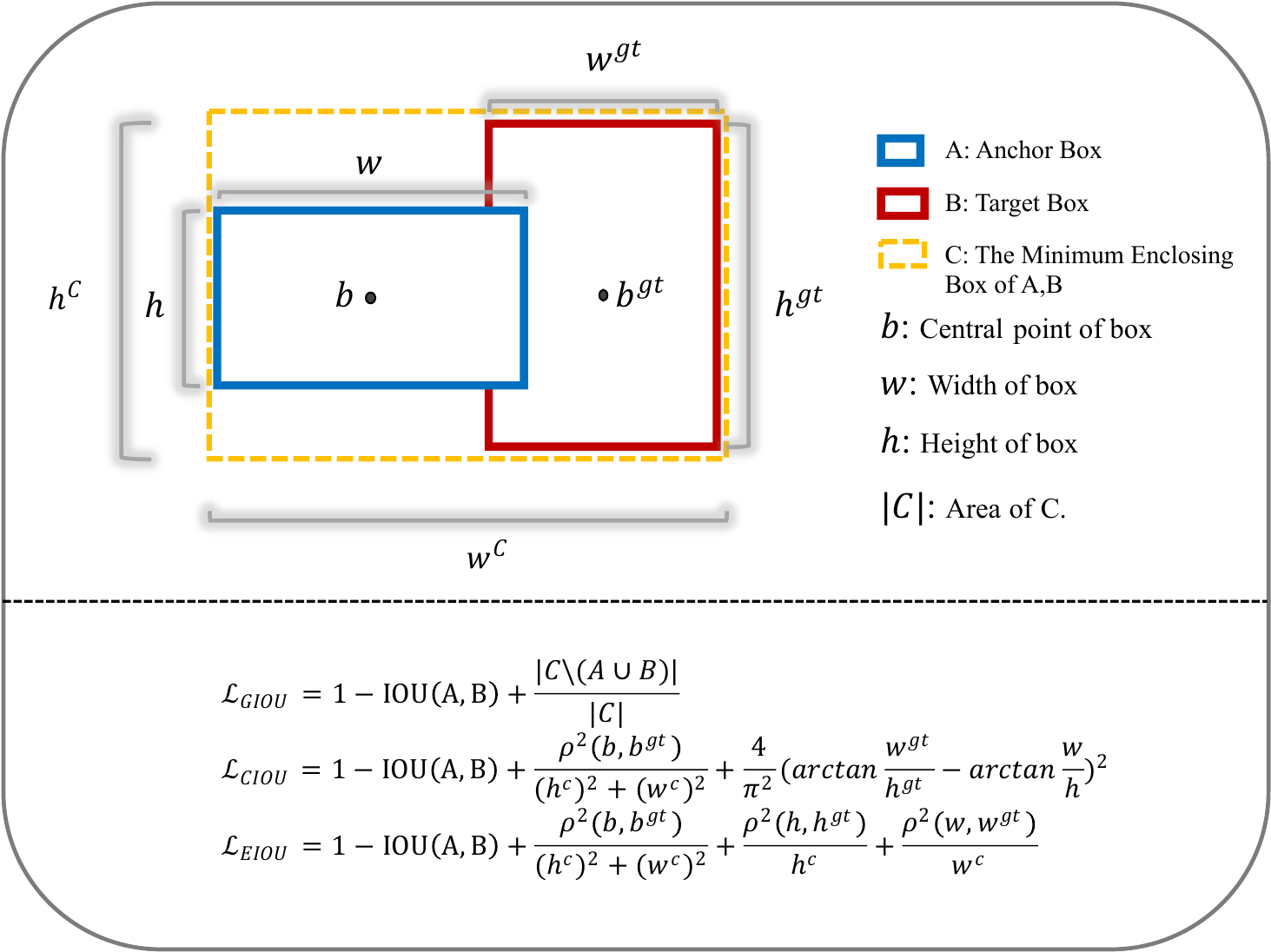}
    \caption{\markchanged{Description of the EIOU loss. The EIOU loss directly minimizes the normalized difference of the target box's and anchor box's width ($w,w^{gt}$), height ($h,h^{gt}$), and central location ($b,b^{gt}$). }}
    \label{fig:eiou_description}
  \end{center}
\end{figure}

\begin{figure}
\centering
\subfigure[]{
  \begin{minipage}[t]{0.45\linewidth}
  \centering
  \includegraphics[scale=0.4]{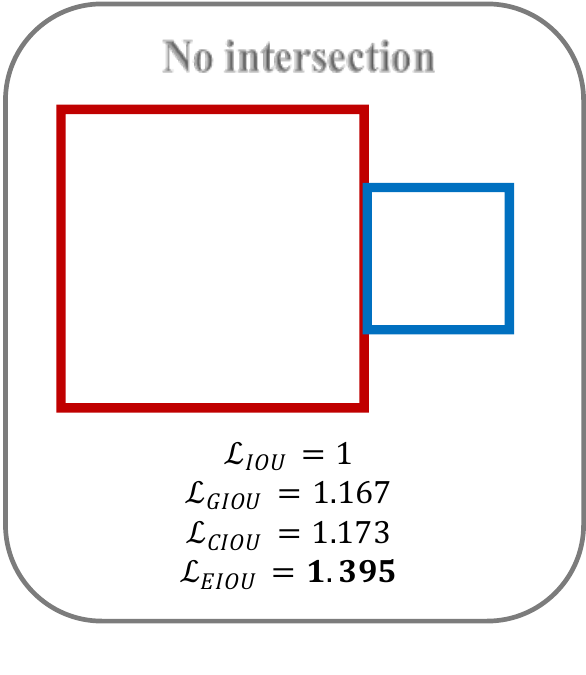}\label{fig:case1}
   \end{minipage}}
   \hfill
   \subfigure[]{
   \begin{minipage}[t]{0.5\linewidth}
   \centering
  \includegraphics[scale=0.4]{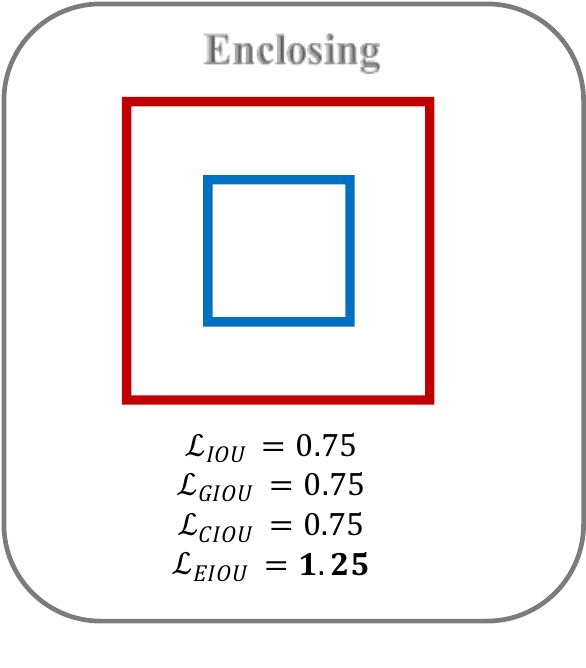}\label{fig:case2}
   \end{minipage}}
  \caption{\markchanged{Flaws of previous IOU-based losses. (a) When two boxes have no intersection. (b) When one box is completely enclosed by another one.}}
\end{figure}

\markchanged{\textbf{Remark.} The calculation of EIOU loss and previous IOU, GIOU, and CIOU loss are described in~\figurename~\ref{fig:eiou_description} and we show the benefit of EIOU loss by analyzing two cases. In~\figurename~\ref{fig:case1}, when two boxes $A,B$ are not intersected and have similar aspect ratio, namely $\frac{w}{h}\approx \frac{w^{gt}}{h^{gt}}$, $IOU(A,B)=0$, where the spatial configuration (position and shape) of the anchor box cannot provide any contribution to the IOU loss. Thus, in such a case, there are not any gradients to optimize the anchor box. GIOU and CIOU loss is a little better than IOU, in contrast, the proposed EIOU loss provides much larger gradients and converges to the target faster. Similarly, when one of these two boxes is enclosed by another (\figurename~\ref{fig:case2}), both GIOU and CIOU loss will degrade to IOU loss and attain slow convergence speed. However, the proposed EIOU loss directly minimizes the difference between the target box’s and anchor box’s width and height but not the aspect ratio as CIOU loss, the EIOU loss can still provide larger gradients in this case.}
\section{Focal Loss For BBR}
In BBR, the problem of imbalanced training examples also exists, i.e., the number of high-quality examples (anchor boxes) with small regression errors is much fewer than low-quality examples (outliers) due to the sparsity of target objects in images. Recent work~\cite{Libra} has shown that the outliers will produce excessively large gradients that are harmful to the training process. Thus, it is of vital importance that making the high-quality examples contribute more gradients to the network training process. As introduced in Section~\ref{sec:EEM}, recent studies~\cite{Libra,DynamicRCNN} have attempted to solve the above problem based on the SmoothL1 loss. In this section, we also start with the SmoothL1 loss and propose FocalL1 loss to increase the contribution of high-quality examples. Furthermore, we find that the simple method cannot be adapted to the IOU-based losses directly. Hence, we finally propose Focal-EIOU loss to improve the performance of the EIOU loss.

\subsection{FocalL1 Loss}

\begin{figure}[t]
\centering
\subfigure[]{
\includegraphics[width=0.24\textwidth]{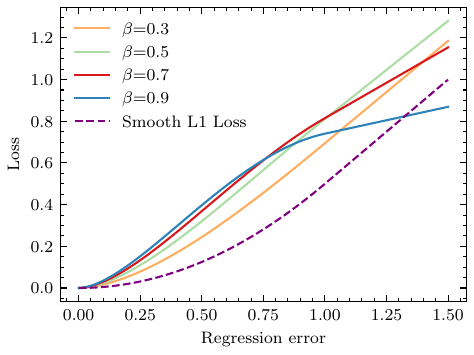}\label{fig:Focal}
}%
\subfigure[]{
\includegraphics[width=0.24\textwidth]{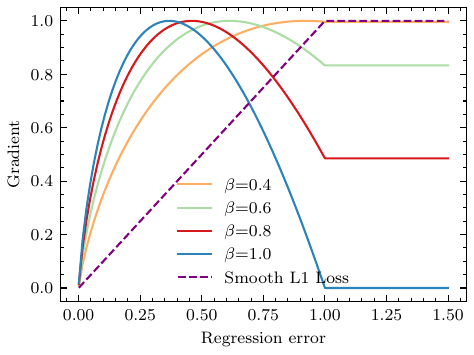}\label{fig:Focal_grad}
}

\caption{Curves for (a) loss and (b) gradient of our FocalL1 loss for bounding box regression. }
\end{figure}

\begin{figure}
\centering
\subfigure[]{
  \begin{minipage}[t]{0.45\linewidth}
  \centering
  \includegraphics[scale=0.5]{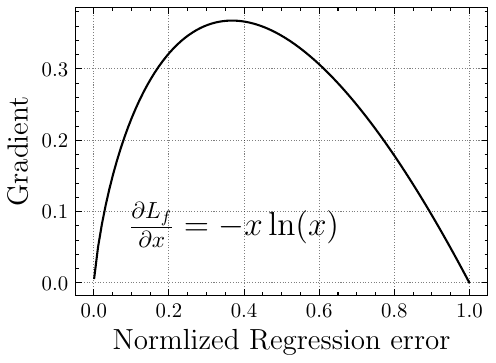}\label{fig:grad_expected}
   \end{minipage}}
   \hfill
   \subfigure[]{
   \begin{minipage}[t]{0.5\linewidth}
   \centering
  \includegraphics[scale=0.5]{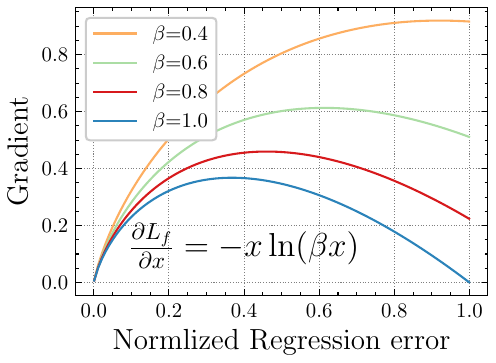}\label{fig:grad_expected_unnorm}
   \end{minipage}}
  \caption{Possible gradient curves. (a) The gradient curve that we expect. (b) Use $\beta$ to control the curves' shape.}
\end{figure}

Firstly, we list the properties of the desirable loss function as follows.
  \begin{enumerate}
    \item When the regression error goes to zero, the gradient magnitude should have a limit of zero.\label{condition1}
    \item The gradient magnitude should increase rapidly around small regression errors and decrease gradually in the area of large regression errors.\label{condition2}
    \item There should be some hyper-parameters to control the degree of inhibition of low-quality examples flexibly.
    \item With variant values of hyper-parameters, the family of gradient functions should have a normalized scale, e.g., $(0, 1]$, which facilitates the balancing between high-quality and low-quality examples.\label{condition4}
  \end{enumerate}
According to the above conditions, as the change of the regression error of bounding box, we can assume an expected function curve of gradient magnitude, which is shown in \figurename~\ref{fig:grad_expected}. The function is $-x\ln x$, satisfying the properties~\ref{condition1} and~\ref{condition2}. Next, we construct a function family with a parameter $\beta$ to control the shape of curves as shown in \figurename~\ref{fig:grad_expected_unnorm}. As $\beta$ increases, the gradient magnitudes of outliers will be further suppressed. However, the gradient magnitudes of high-quality examples will also decrease, which is not what we expect. Thus, we add another parameter $\alpha$ to normalize the gradient magnitudes with different $\beta$ into $[0,1]$ as required by property ~\ref{condition4}. Finally, the family of gradient magnitude functions can be formulated as follows.
\begin{equation}\label{equ:grad_focal}
g(x)=\frac{\partial L_f}{\partial x}=\left\{
        \begin{array}{ll}
          -\alpha x\ln(\beta x), & 0< x\leq 1; 1/e\leq\beta\leq1,   \\
          -\alpha \ln(\beta ), & x>1; 1/e\leq\beta\leq1.
        \end{array}
      \right.
\end{equation}

Here, the value range of $\beta$ is obtained due to the following reasons.. When $x\in(0,1]$, $g''(x)=-\frac{\alpha}{x}\leq 0$, which means $g(x)$ is a concave function with a global maximal value. Solving $g'(x)=0$, we can get $x^*=\frac{1}{e\beta}$. As $x^*\in(0,1]$, $\frac{1}{e\beta}\in[0,1]\rightarrow \beta\in[1/e,\infty]$. We also must ensure $\beta x\in(0,1]$, then, $\beta\in[1/e,1]$. To satisfy the property ~\ref{condition4}, we set the maximal value $f(x^*)=1$ and get the relation between $\alpha$ and $\beta$: $\alpha=e\beta$.
By integrating the gradient formulation above, we can get the FocalL1 loss for BBR,
\begin{equation}\label{equ:BBR_focal}
L_f(x)=\left\{
        \begin{array}{ll}
          -\frac{\alpha x^2(2\ln (\beta x)-1)}{4}, & 0< x\leq 1; 1/e\leq\beta\leq1,   \\
          -\alpha \ln(\beta)x + C,& x>1; 1/e\leq\beta\leq1,
        \end{array}
      \right.
\end{equation}
where $C$ is a constant value. To ensure $L_f$ in Eq.~\eqref{equ:BBR_focal} is continuous at $x=1$, we have $C=(2\alpha\ln\beta+\alpha)/4$.


\figurename~\ref{fig:Focal_grad} shows the proposed FocalL1 loss can increase the value of inliers' gradients and suppress the value of outliers' gradients according to $\beta$. A larger $\beta$ requires the inliers to have few regression errors and quickly suppresses the gradients'value of outliers. Similarly, in \figurename~\ref{fig:Focal} the blue curve denotes the maximal value of $\beta$. With the increase of regression error, the loss of the blue curve first increases rapidly and then tends to be stable. The orange curve with the minimal $\beta$ value is growing faster and faster, reaching its peak around $x=1$.
Now we can calculate the localization loss by FocalL1 loss, $L_{Loc}=\sum_{i\in\{x,y,w,h\}}L_f(|B_i-B_i^{gt}|)$, where ${B}$ is the regression result and ${B^{gt}}$ is the regression target.
\subsection{Focal-EIOU Loss}
To enable the EIOU loss focus on high-quality examples, one can naturally consider replacing $x$ in Eq.~\eqref{equ:BBR_focal} with the EIOU loss. However, we observe that the above combination doesn't work well. The analysis is as follows. 

Given the offset $\ell_1(B_i)=|B_i-B_i^{gt}|$, the gradient of the FocalL1 loss is $\frac{\partial L_f(\ell_1(B_i))}{\partial B_i}=\frac{\partial L_f}{\partial \ell_1}\frac{\partial \ell_1}{\partial B_i}$, where ${\partial \ell_1}/{\partial B_i}$ is a constant equals $1$ or $-1$. Thus, even the offset is small, ${\partial L_f}/{\partial \ell_1}$ can also bring enough gradients to make the model continuously optimized.
However, if we replace the offset $\ell_1(B_i)$ with $L_{EIOU}({B,B^{gt}})$, the gradient can be calculated as $\frac{\partial L_f}{\partial L_{EIOU}}\frac{\partial L_{EIOU}}{\partial B_i}$.
Here, the $\partial {L_{EIOU}}/\partial_{B_i}$ is not a constant value any more. Moreover, it will be very small in our empirical studies as the ${L_{EIOU}}$ approaches to zero, while the $\partial L_f/\partial L_{EIOU}$ is also near to zero at that time. Thus, after the multiplication, the overall gradient will be even more smaller, which weakens the effect of reweighting on the boxes with small ${L_{EIOU}}$. To tackle this problem, we use the value of IOU to reweight the EIOU loss and get Focal-EIOU loss as follows


\begin{equation}\label{equ:Focal-EIOU}
  L_{\text{Focal-EIOU}} = IOU^\gamma L_{EIOU},
\end{equation}
where $IOU={|A\cap B|}/{|A\cup B|}$ and $\gamma$ is a parameter to control the degree of inhibition of outliers. We also try other forms of the reweighting process in Section~\ref{section:ExpFocal-EIOU}, while we find Eq.~\eqref{equ:Focal-EIOU} achieves superior performance.

\section{Experiments}

\subsection{Datasets and Evaluation Metrics}
  \begin{figure}
\centering
\subfigure[]{
  \begin{minipage}[t]{0.45\linewidth}
  \centering
  \includegraphics[scale=0.45]{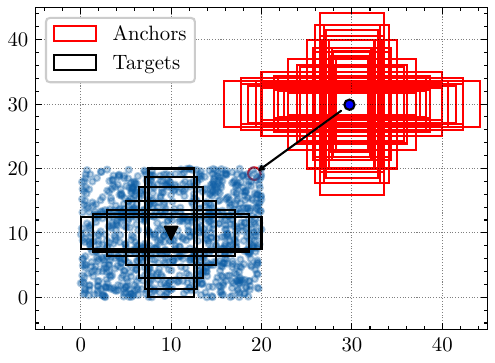}\label{fig:sim}
   \end{minipage}}
   \hfill
   \subfigure[]{
   \begin{minipage}[t]{0.45\linewidth}
   \centering
  \includegraphics[scale=0.45]{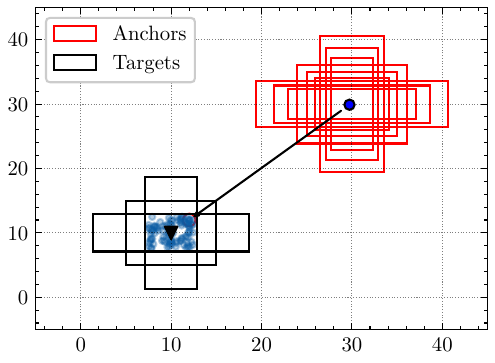}\label{fig:sim2}
   \end{minipage}}
  \caption{Simulation setup: (a) Setup $1$: $7$ target box and $1000\times 7\times 7$ anchors. (b) Setup $2$: $3$ target box and $100\times 3\times 3$ anchors.}

\end{figure}
\begin{algorithm}[t]
  \caption{Simulation Experiments}
  \begin{algorithmic}[1]
  \Require
  $A=7\text{ or }3$ represents simulation setup $1$ or setup $2$. $\{\{B_{n,s}\}_{s=1}^S\}\}_{n=1}^{N}$ denotes all the anchors at $N$ points, where $S=A*A$ is the number of combinations of different areas and aspect ratios. $\{B_i^{gt}\}_{i=1}^{A}$ is the set of target boxes in $(10,10)$ with an area of $100$.
   \Ensure
  IOU $\mathbf{I}\in\mathcal{R}^{T\times N\times S\times A}$ of each target-anchor box pair and regression error $\mathbf{E}\in\mathcal{R}^{T}$ in each iteration, where $T$ is the maximal iteration and $N$ is the number of generated points.
   \State $(T,\mathbf{E},\mathbf{I})\leftarrow (200,\mathbf{0},\mathbf{0})$
   \For {$t=1$ to $T$}
    \For {$n=1$ to $N$}
        \For {$s=1$ to $S$}
          \For {$i=1$ to $A$}
              \If { $t\leq 0.8T$} $\mu = 0.1$
              \ElsIf {$t\leq 0.9T$} $\mu=0.01$
              \Else $\;\mu=0.001$
              \EndIf
              \State $\nabla B_{n,s}^{t-1}={\partial L( B_{n,s}^{t-1},B_i^{gt})}/{\partial B_{n,s}^{t-1}}$
              \State $B_{n,s}^t=B_{n,s}^{t-1}+\mu\nabla B_{n,s}^{t-1}$
              \State $\mathbf{E}(t)=\mathbf{E}(t)+|B_{n,s}^t-B_{i}^{gt}|$
              \State $\mathbf{I}(t,n,s,i)=IOU(B_{n,s}^t,B_{i}^{gt})$
   \EndFor
   \EndFor
   \EndFor
   \EndFor
   \State \textbf{Return} $\mathbf{E,I}$
  \end{algorithmic}
  \label{algorithm:sim}
\end{algorithm}
We conduct experiments with both synthetic and real datasets. For the synthetic data, we conduct two simulation experiments to respectively study the superiorities of the EIOU and Focal-EIOU loss, as well as the importance of EEM. As shown in ~\figurename~\ref{fig:sim} , we randomly generate $1000$ points within a $20\times 20$ box.  Each point has a set of $7\times 7$ anchors with different aspect ratios ( $1$:$4$, $1$:$3$, $1$:$2$, $1$:$1$, $2$:$1$, $3$:$1$ and $4$:$1$) and different scales ($50$, $67$, $75$, $100$, $133$, $150$ and $200$). {The blue points denote the central points of anchors, where the red rectangles illustrate the $7\times 7$ scales of the anchors around one of the 1000 blue central points}. We also have $7$ target boxes locating at ($10$,$10$). These target boxes all have a fixed scale $100$ but different aspect ratios ( $1$:$4$, $1$:$3$, $1$:$2$, $1$:$1$, $2$:$1$, $3$:$1$ and $4$:$1$). To emphasize the importance of EEM, we initialize higher quality anchors than those in~\figurename~\ref{fig:sim}. { We further constrain to sample  anchors within a smaller square (the blue area) around the target locations, so that the anchors will have heavy overlaps with targets, which simulate the common situations in real scenarios.} As depicted in ~\figurename~\ref{fig:sim2}, $100$ points are generated within a square whose center point is ($10$,$10$) and the side length is $2.5$. Each point has a set of $3\times 3$ anchors with different aspect ratios ($1$:$3$, $1$:$1$, $3$:$1$) and different scales ($50$, $100$, $150$). Targets boxes have three aspect ratios ($1$:$3$, $1$:$1$, $3$:$1$) and a fixed scale $100$.  The simulation algorithm is depicted in Algorithm~\ref{algorithm:sim}. We go through all the anchors and regress them to each target. For specific anchor $B_{n,s}$ and target $B_i^{gt}$, we regress anchor $B_{n,s}^{t-1}$ to $B_{n,s}^{t}$ according to the gradient of the loss $L$ w.r.t $B_{n,s}^{t-1}$ at iteration $t$. The performance of the regression process is evaluated with $\ell_1$ loss and IOU metric. 

We also present the experimental results on the bounding box detection track of the challenging COCO 2017 dataset~\cite{coco}. We use the COCO train-2017 split ($115k$ images) for training and report the ablation studies on the val-2017 split ($5k$ images). The COCO-style Average Precision (AP) is chosen as the main evaluation metric. 

\subsection{Implementation Details}

For fair comparisons, all experiments are implemented with PyTorch~\cite{pytorch}. The backbones used in the experiments are publicly available. For most experiments on COCO 2017 dataset, we use ResNet-50 backbone and run $90k$ iterations. For comparisons with previous methods we run $90k$ iterations with various backbones. We train detectors with $4$ GPUs ($4$ images per GPU), adopting the stochastic gradient descent (SGD) optimizer with the initial learning rate $0.01$ and decaying it by a factor of $0.1$ at $50k,70k$ and $80k$. We doesn't following previous settings that decay the learning rate twice because EIOU loss brings larger gradients and high learning rate leads to unstable training. The default weight for BBR is set to $1.0$ for $\ell_n$-norm losses and $2.5$ for IOU-based losses. All other hyper-parameters follow the settings in ATSS~\cite{paa} if not specifically noted.

In order to avoid the slow convergence speed in the early training period due to reweighting, we use the sum of weights in each batch to normalize the Focal-EIOU loss. Formally,

\begin{equation}\label{equ:FocalNorm}
  {L}_{\text{Focal-EIOU}}=\frac{\sum_{i=1}^nW_i\cdot L_{EIOU_i}}{\sum_{i=1}^n W_i},
\end{equation}
\noindent where $n$ is the number of anchor-target pairs in each batch. $L_{EIOU_i}$ and $W_i$ are the EIOU loss of anchor-target pair $i$ and the corresponding weight.

\subsection{Simulation Experiments}\label{sec:sim}

\begin{figure}
  \centering
  \subfigure[]{
    \begin{minipage}[t]{0.5\linewidth}
    \centering
    \includegraphics[scale=0.25]{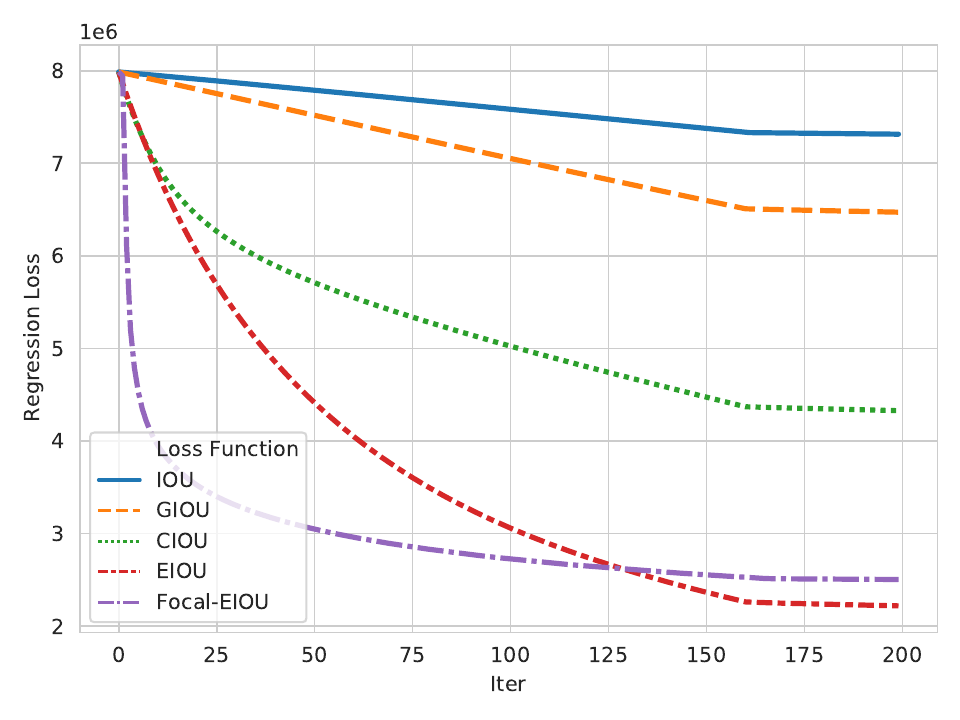}\label{fig:sim_result1L1}
     \end{minipage}}
     \hfill
     \subfigure[]{
     \begin{minipage}[t]{0.45\linewidth}
     \centering
    \includegraphics[scale=0.27]{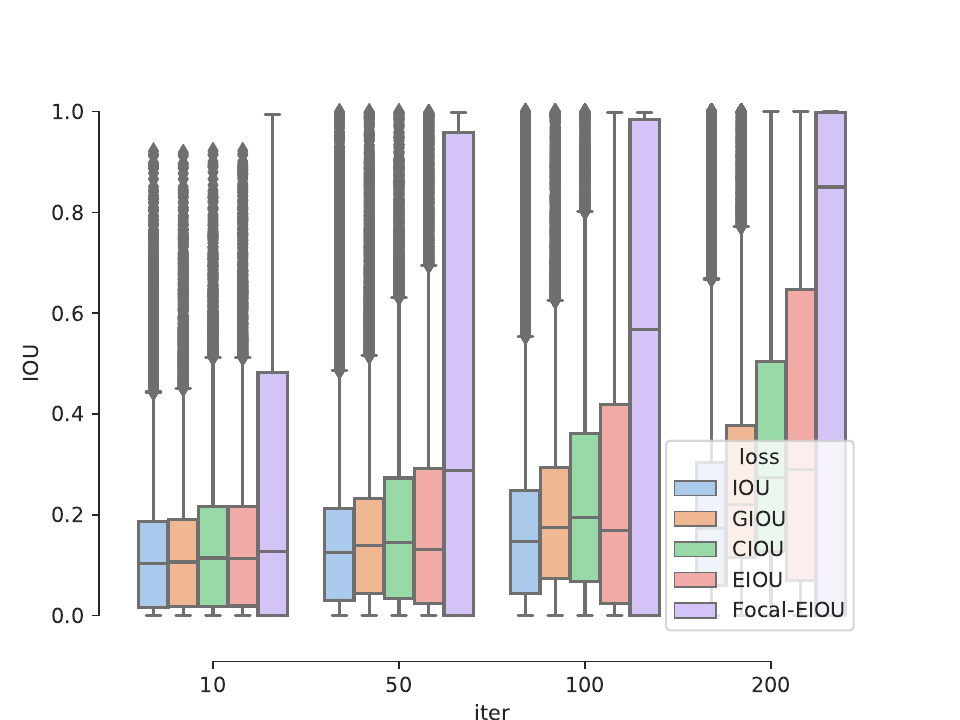}\label{fig:sim_result1IOU}
     \end{minipage}}
     \caption{Simulation experimental results $1$: (a) Regression error sum curves of different loss functions. (2) Variation trend of box plot of IOU with different loss functions. The Focal-EIOU loss with parameters $\gamma=0.5$.}
     \label{fig:sim_result1}
  \end{figure}



    \figurename~\ref{fig:sim_result1L1} and \figurename~\ref{fig:sim_result1IOU} show the simulation results when most of the anchors are low quality examples. It verifies that the EIOU loss has a faster convergence speed and better regression accuracy than the IOU, GIOU and CIOU losses. Note that, the regression error of Focal-EIOU loss is larger than the EIOU loss, which is depicted in~\figurename~\ref{fig:sim_result1L1}. While in ~\figurename~\ref{fig:sim_result1IOU}, the Focal-EIOU loss is less concerned about hard examples and these examples bring lots of regression errors for the Focal-EIOU loss. However, both the number and quality of high-quality examples of the Focal-EIOU loss are much higher than the other four kinds of loss functions. 
    
\begin{figure}
    \centering
    \subfigure[]{
      \begin{minipage}[t]{0.5\linewidth}
      \centering
      \includegraphics[scale=0.25]{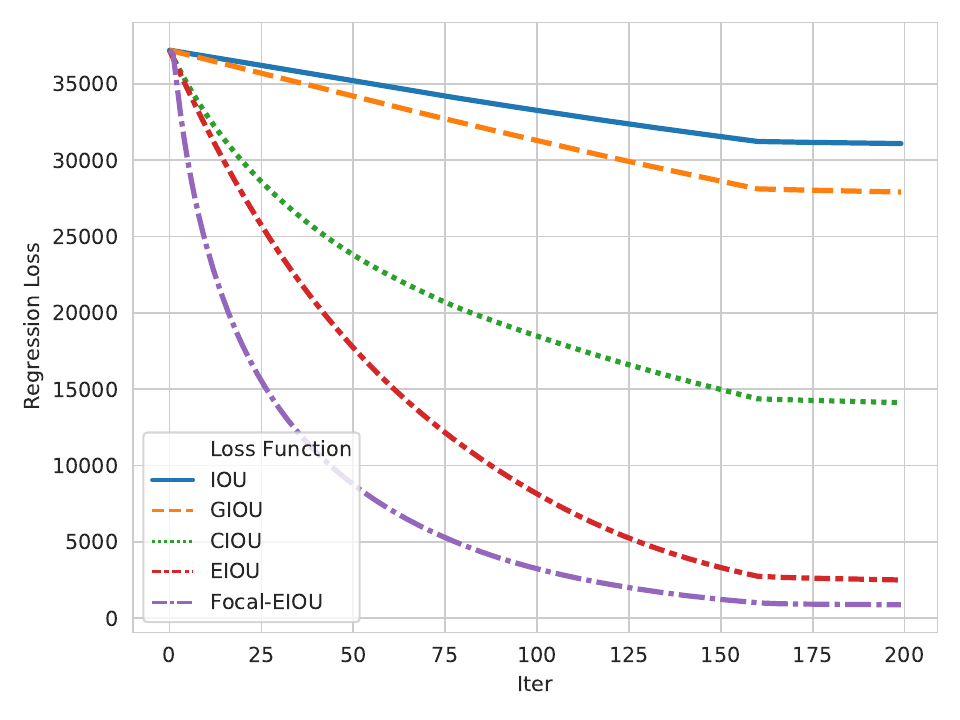}\label{fig:sim_result2L1}
      \end{minipage}}
      \hfill
      \subfigure[]{
      \begin{minipage}[t]{0.45\linewidth}
      \centering
      \includegraphics[scale=0.27]{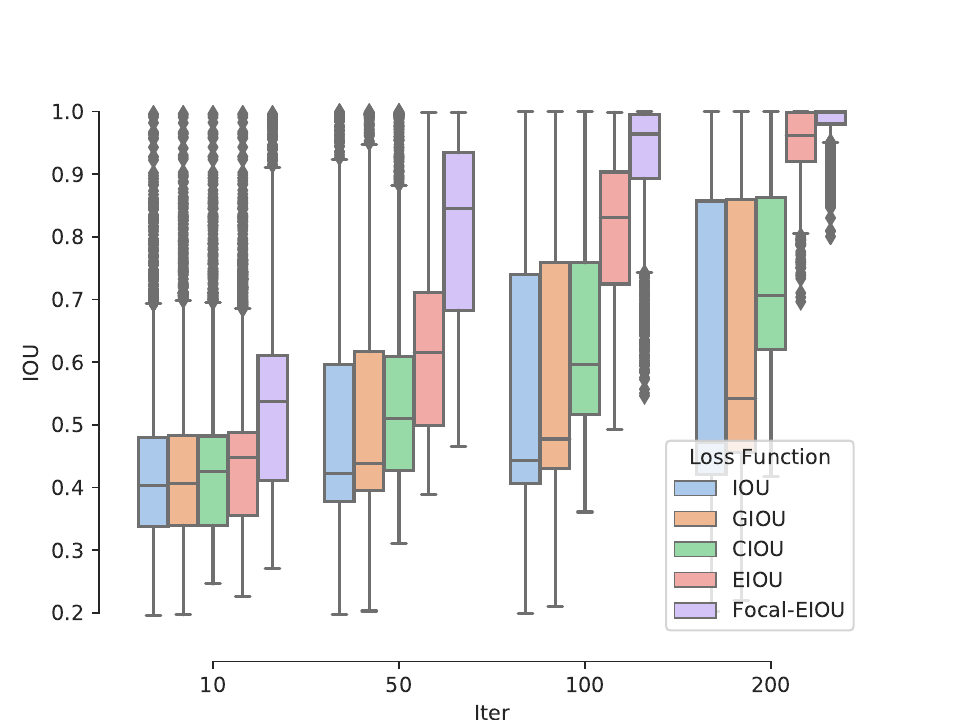}\label{fig:sim_result2IOU}
      \end{minipage}}
      \caption{Simulation experimental results $2$: (a) Regression error sum curves of different loss functions. (2) Variation trend of box plot of IOU with different loss functions. The Focal-EIOU loss with parameters $\gamma=0.5$.}
      \label{fig:sim_result2}
    \end{figure}

    \figurename~\ref{fig:sim_result2L1} and \figurename~\ref{fig:sim_result2IOU} demonstrates the importance of EEM. The Focal-EIOU loss shows its extraordinary dominance over other IOU based losses. It not only has the fastest convergence speed and lowest regression error (\figurename~\ref{fig:sim_result2L1}), but also improves the quality of these high-quality examples at a speed far beyond other loss functions.
    Compared with the EIOU loss, the Focal-EIOU loss has a longer tail, while the mean value of IOU in \figurename~\ref{fig:sim_result2IOU} is much higher. In other words, although the Focal-EIOU loss has some low-quality regressed anchors, it indeed has much more high-quality regressed anchors than the EIOU loss. 

    From the above simulation experiments, both the EIOU and the Focal-EIOU loss have achieved faster convergence speed. The Focal-EIOU loss has lower localization errors due to the reweighting on those high-quality examples.
\subsection{Ablation Experiments}
 \begin{table}[h]
      \begin{tabular}{c c c c c c c}
        \toprule[1pt]
        Method & AP & $\text{AP}_{50}$ & $\text{AP}_{75}$ & $\text{AP}_{S}$ & $\text{AP}_{M}$ & $\text{AP}_{L}$ \\ \hline
        Baseline &35.9 & 55.2 & 38.4 & 21.2 & 39.5 & 48.4\\\hline
        IOU & 36.5 & 55.6 & 38.9 & 20.9 & 40.1 & 48.0 \\
        GIOU & 36.5 & 55.6 & 39.0 & 20.7 & 40.2 & 48.2 \\
        CIOU & 36.7 & 55.7 & 39.2 & 20.6 & 40.4 & 49.0 \\\hline
        FocalL1 & 36.5 & 55.8 & 38.9 & $\mathbf{21.2}$ & 39.8 & 48.8\\
        EIOU & 37.0 & 55.7 & 39.5 & 20.7 & 40.5 & 49.5\\  
        Focal-EIOU (v1) & 36.8 & 55.4 & 39.5 & 20.9 & 40.0 & 49.1\\
        Focal-EIOU & $\mathbf{37.5}$ & $\mathbf{56.1}$ & $\mathbf{40.0}$ & 21.1 & $\mathbf{40.9}$ & $\mathbf{49.8}$\\
        \bottomrule[1pt]\\
      \end{tabular}
      \centering
      \caption{    Overall ablation studies on COCO val-2017. }
      \centering
      \label{tab:overall}
      \end{table}

\subsubsection{Overall Ablation Studies}

To demonstrate the effectiveness of each proposed component, we report the overall ablation studies in Table~\ref{tab:overall}. The FocalL1 loss improves the box AP from $35.9\%$ to $36.5\%$. The EIOU loss brings $1.1\%$ higher box AP than the ResNet-50 FPN RetinaNet baseline. Directly applying Eq.~\eqref{equ:BBR_focal} to the EIOU loss (namely use the EIOU loss as $x$ in Eq.~\eqref{equ:BBR_focal}), we can get Focal-EIOU (v1) and it doesn't work well. However, the Focal-EIOU loss in Eq.~\eqref{equ:Focal-EIOU} brings reasonable gains, improving the baseline's AP by $1.6\%$. We also conduct experiments with the IOU, GIOU, CIOU and EIOU losses in Table~\ref{tab:overall}. For fair comparisons, we set the weight for BBR to $2.5$. Experimental results show that the performance of other IOU-based losses is inferior to the proposed method. Then we conduct detailed ablations for each block and hyper-parameter.
\begin{figure*}
  \centering
  \subfigure[]{
    \begin{minipage}[t]{0.3\linewidth}
    \centering
    \includegraphics[scale=0.3]{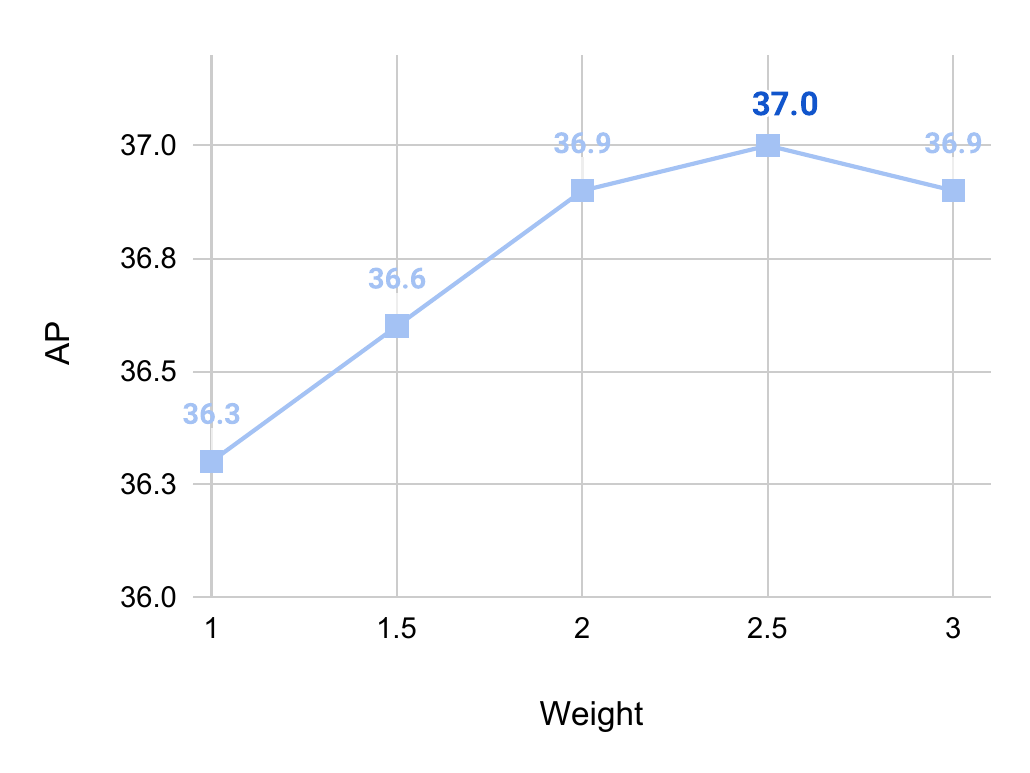}\label{fig:weight1}
     \end{minipage}}
     \hfill
     \subfigure[]{
     \begin{minipage}[t]{0.3\linewidth}
     \centering
    \includegraphics[scale=0.3]{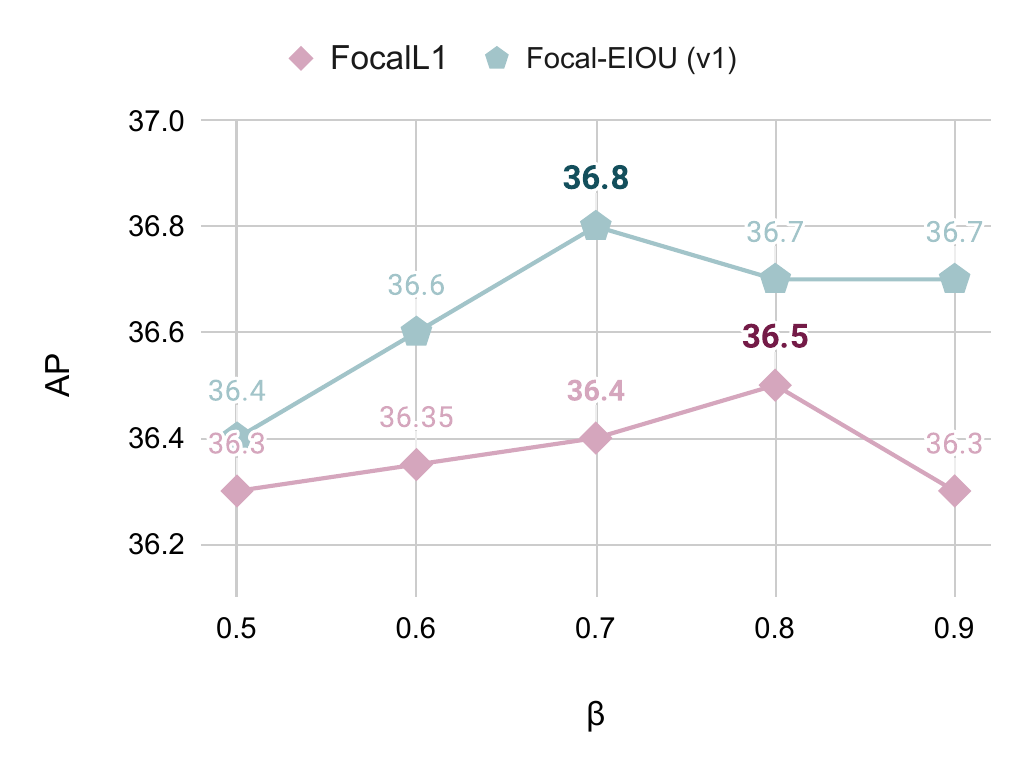}\label{fig:weight2}
     \end{minipage}}
      \hfill
      \subfigure[]{
      \begin{minipage}[t]{0.3\linewidth}
      \centering
      \includegraphics[scale=0.3]{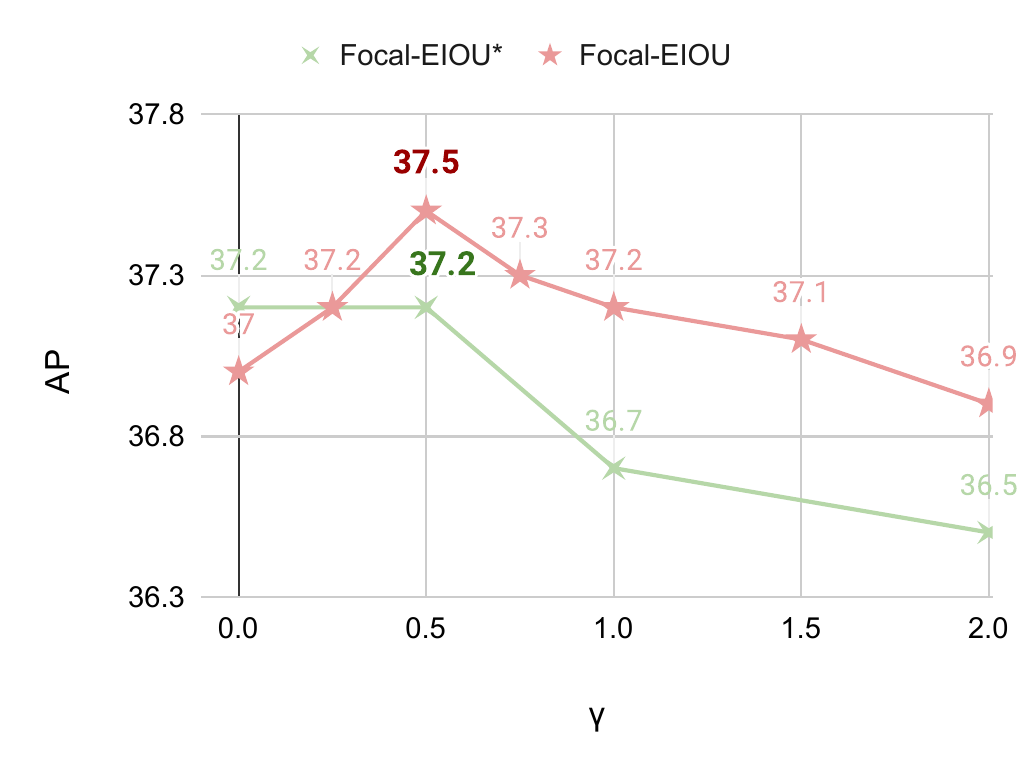}\label{fig:weight4}
      \end{minipage}}
     \caption{Performance of methods with different values of parameters. (a) The EIOU loss with different weights for BBR.  (b) The FocalL1 and Focal-EIOU (v1) loss with different $\beta$. (c) The Focal-EIOU and Focal-EIOU* loss with different $\gamma$.}
  \end{figure*}
\subsubsection{Effect of the Tradeoff Weight}
 We first verify the effects of the weight for BBR, where the weight here controls the balance between the classification loss and BBR loss in object detection. \figurename~\ref{fig:weight1} shows that tuning the weight can improve the performance with gains of $1.1\%$ AP. The performance with a loss weight larger than $2.5$ starts to drop down. These results indicate that these outliers have a negative impact on the training process and we do not fully utilize the potential of the model architecture.
 \begin{table}[H]
  \begin{tabular}{c c c c c c c}
    \toprule[1pt]
    Setting & AP & $\text{AP}_{50}$ & $\text{AP}_{75}$ & $\text{AP}_{S}$ & $\text{AP}_{M}$ & $\text{AP}_{L}$ \\ \hline
    Baseline & 35.9 & 55.5 & 38.4 & 21.2 & 39.5 & 48.4\\\hline
  
    BalancedL1 & 36.3 & 55.3 & 38.8 & 20.5 & 39.3 & 48.5 \\
    FocalL1 &  $\mathbf{36.5}$ & $\mathbf{55.8}$ & $\mathbf{38.9}$ & $\mathbf{21.2}$ & $\mathbf{39.8}$ & $\mathbf{48.8}$ \\
    \bottomrule[1pt]\\
  \end{tabular}
  \centering
  \caption{    Ablation studies of the FocalL1 loss on COCO val-2017. }
  \centering
  \label{tab:bbrFocal}
  \end{table}

\subsubsection{Ablation Studies on FocalL1 Loss}

      \begin{table*}[]
      \centering
  \adjustbox{max width=\textwidth}{%
        \begin{tabular}{c c c c c c c c c}
          \toprule[1pt]
          Method & Backbone & AP & $\text{AP}_{50}$ & $\text{AP}_{75}$ & $\text{AP}_{S}$ & $\text{AP}_{M}$ & $\text{AP}_{L}$ \\ \hline
          Faster R-CNN~\cite{faster} & ResNet-50-FPN &37.3 & 58.2 & 40.3 & $\mathbf{21.3}$ & 40.9 & 48.0\\
          Faster R-CNN* & ResNet-50-FPN &$\mathbf{38.9}$ & $\mathbf{59.1}$ & $\mathbf{42.4}$ & 21.2 & $\mathbf{41.1}$ & $\mathbf{50.2}$\\ \hline
          Faster R-CNN~\cite{faster} & ResNeXt-101-32x4d-FPN &41.2 & 62.1 & 45.1 & 24 & 45.5 & 53.5\\
          Faster R-CNN* & ResNeXt-101-32x4d-FPN &$\mathbf{42.4}$ & $\mathbf{63.1}$ & $\mathbf{46.8}$ & $\mathbf{24.2}$ & $\mathbf{46.3}$ & $\mathbf{54.1}$\\ \hline
          Mask R-CNN~\cite{maskrcnn} & ResNet-50-FPN &38.2 & 58.8 & 41.4 & 21.9 & 40.9 & 49.5\\
          Mask R-CNN* & ResNet-50-FPN & $\mathbf{39.6}$ & $\mathbf{59.3}$ & $\mathbf{41.7}$ & $\mathbf{22.4}$ & $\mathbf{41.5}$ & $\mathbf{51.1}$\\ \hline
          Mask R-CNN~\cite{maskrcnn} & ResNeXt-101-32x4d-FPN &41.9 & 62.5 & 45.9 & 24.4 & 46.3 & 54.0\\
          Mask R-CNN* & ResNeXt-101-32x4d-FPN & $\mathbf{43.0}$ & $\mathbf{63.1}$ & $\mathbf{46.1}$ & $\mathbf{24.4}$ & $\mathbf{47.3}$ & $\mathbf{56.1}$\\ \hline 
          RetinaNet~\cite{fatal_loss} & ResNet-50-FPN &35.9 & 55.2 & 38.4 & 21.2 & 39.5 & 48.4\\
          RetinaNet* & ResNet-50-FPN &$\mathbf{37.5}$ & $\mathbf{56.1}$ & $\mathbf{40.0}$ & $\mathbf{21.3}$ & $\mathbf{40.9}$ & $\mathbf{49.8}$\\ \hline
          RetinaNet~\cite{fatal_loss} & ResNeXt-101-32x4d-FPN	 &40.8 & 60.9 & 43.7 & $\mathbf{22.9}$ & 44.5 & 54.6\\
          RetinaNet* & ResNeXt-101-32x4d-FPN &$\mathbf{41.8}$ & $\mathbf{61.4}$ & $\mathbf{44.7}$ & 21.7 & $\mathbf{45.0}$ & $\mathbf{55.2}$\\ \hline
          ATSS~\cite{ATSS} & ResNet-50-FPN &39.1 & 57.6 & 42.1 & $\mathbf{22.9}$ & 42.8 & 51.1\\
           ATSS* & ResNet-50-FPN &$\mathbf{39.7}$ & $\mathbf{57.9}$ & $\mathbf{45.7}$ & 22.6 & $\mathbf{43.2}$ & $\mathbf{51.8}$\\ \hline
          ATSS & ResNeXt-64x4d-101-DCN &50.7 & 68.9 & 56.3 & $\mathbf{33.2}$ & 52.9 & 62.2\\
           ATSS* & ResNeXt-64x4d-101-DCN &$\mathbf{51.4}$ & $\mathbf{69.3}$ & $\mathbf{57.3}$ & 32.8 & $\mathbf{53.4}$ & $\mathbf{64.2}$\\ \hline
           PAA~\cite{paa} & ResNet-50-FPN &40.3 & 57.6 & 43.9 & $\mathbf{23.0}$ & 44.9 & 54.0\\
           PAA* & ResNet-50-FPN  & $\mathbf{40.8}$ & $\mathbf{57.9}$ & $\mathbf{44.7}$ & 22.9 & $\mathbf{45.3}$ & $\mathbf{54.9}$\\ \hline
           PAA~\cite{paa} & ResNeXt-64x4d-101-DCN &51.4 & 69.7 & 57.0 & 34.0 & 53.8 & 64.0\\
           PAA* & ResNeXt-64x4d-101-DCN  & $\mathbf{52.3}$ & $\mathbf{70.2}$ & $\mathbf{58.1}$ & 33.2 & $\mathbf{54.3}$ & $\mathbf{66.2}$\\ \hline
          DETR~\cite{carion2020end} & ResNet-50 &42.0 & 62.4 & 44.2 & 20.5 & 45.8 & 61.1\\\Red
          DETR* & ResNet-50  & $\mathbf{43.2}$ & $\mathbf{62.5}$ & $\mathbf{45.3}$ & \textbf{21.1} & $\mathbf{46.1}$ & $\mathbf{62.7}$\\ 
           \bottomrule[1pt]
        \end{tabular}}
        \caption{    The performance when incorporating the Focal-EIOU loss with different SOTA models. * indicates using the Focal-EIOU loss instead of their original losses.}
        \label{tab:sota}
        \end{table*}
As shown in \figurename~\ref{fig:weight2}, we test different $\beta$ of the FocalL1 loss. Generally speaking, setting a larger $\beta$ will further suppress the gradients of low-quality examples, but increase the gradients of high-quality examples. Finally, we find that setting $\beta=0.8$ achieves the best trade-off and $36.5\%$ AP, which is $0.6\%$ higher than the ResNet-50 FPN RetinaNet baseline. We also reimplement the BalancedL1 loss with the superior parameters proposed in the previous work~\cite{Libra} and it brings $0.4\%$ improvement on AP (Table~\ref{tab:bbrFocal}). These experimental results show that the FocalL1 loss makes the model better.

\subsubsection{Ablation Studies on Focal-EIOU Loss.}\label{section:ExpFocal-EIOU}

To illustrate the improvements brought by different methods for reweighting the EIOU loss, we compare three reweighting methods here. We firstly show that the form of the FocalL1 loss is not suitable, namely using the EIOU loss as $x$ in Eq.~\eqref{equ:BBR_focal}. The experimental results are shown in \figurename~\ref{fig:weight2}. As we mentioned before, applying the FocalL1 loss directly to the EIOU loss leads to the reduction of the high-quality examples' gradients, which is not suitable for and thus cannot improve the performance of the EIOU loss.

We then use the focal loss ~\cite{fatal_loss} to reweight the EIOU loss and get the Focal-EIOU* loss, i.e., ${L}_{\text{Focal-EIOU}^*}=-(1-IOU)^\gamma\log(IOU)EIOU$. Originally, the focal loss works well when facing the extreme foreground-background class imbalance. Results in \figurename~\ref{fig:weight4} show that although we can gain performance improvements, it quickly decrease with the increase of $\gamma$. The reason is that we cannot suppress the gradients of hard examples in such an extreme way due to their effectiveness in the BBR process.

  Finally we evaluate the proposed method in Eq.~\eqref{equ:Focal-EIOU}. \figurename~\ref{fig:weight4} shows that compared to the focal loss, the proposed method brings more stable improvement. 
 Larger $\gamma$ brings stronger suppression on hard examples and may retard the convergence speed. This is also the reason why the performance is poorer than the baseline when $\gamma=2.0$. We find that setting $\gamma=0.5$ achieves the best trade-off and use it as the default value for further experiments.



\subsection{Incorporations with State-of-the-Arts}
In this subsection, we evaluate the Focal-EIOU loss by incorporating it into popular object detectors including Faster R-CNN~\cite{faster}, Mask R-CNN~\cite{maskrcnn}, RetinaNet~\cite{fatal_loss}, PAA~\cite{paa} and ATSS~\cite{ATSS}, and {DETR}~\cite{carion2020end}. Results in Table~\ref{tab:sota} show that training these models by the Focal-EIOU loss can consistently improve their performance compared to their own regression losses. Compared to other models, the ATSS and PAA attain relatively small improvements. There are two reasons: (i) The ATSS and PAA both use the GIOU loss with carefully adjusted parameters. These parameters may not be suitable for the proposed Focal-EIOU loss. (ii) Both ATSS and PAA have the reweighting processes on their own loacalization losses ( In PAA, it is the predicted IOUs with corresponding target box. In ATSS, it is the centerness scores when using the centerness prediction). Although these reweighting methods limit the improvement brought by the Focal-EIOU loss, they confirm the necessity of EEM.

Finally, we emphasize that $L_\text{IOU}$, $L_\text{GIOU}$ only achieves $0.1\%$ AP improvement (Table 5 in \cite{2019giou}) and $L_\text{CIOU}$ achieves $0.72 \%$ AP improvement (Table.3 in~\cite{2020diou}) when using Faster R-CNN as the baseline. By comparison, the Focal-EIOU loss brings $1.6 \%$ AP improvement. It is a relatively huge performance gain, verifying our method is simple but efficient. Recent work Generalized Focal Loss~\cite{li2020generalized} consists of two parts: Quality Focal Loss (QFL) and Distribution Focal Loss (DFL). The former is not relevant to the BBR. The latter is an efficient loss for BBR, while it also ignored the importance of EEM. From Table 3 in~\cite{li2020generalized}, we can see that using DFL only improves the performance of ATSS from $39.2\%$ to $39.5\%$ AP while our Focal-EIOU loss brings $0.6\%$ AP improvement.
 \begin{figure*}
  \centering
  \subfigure[]{
    \begin{minipage}[t]{0.3\linewidth}
    \centering
    \includegraphics[scale=0.25]{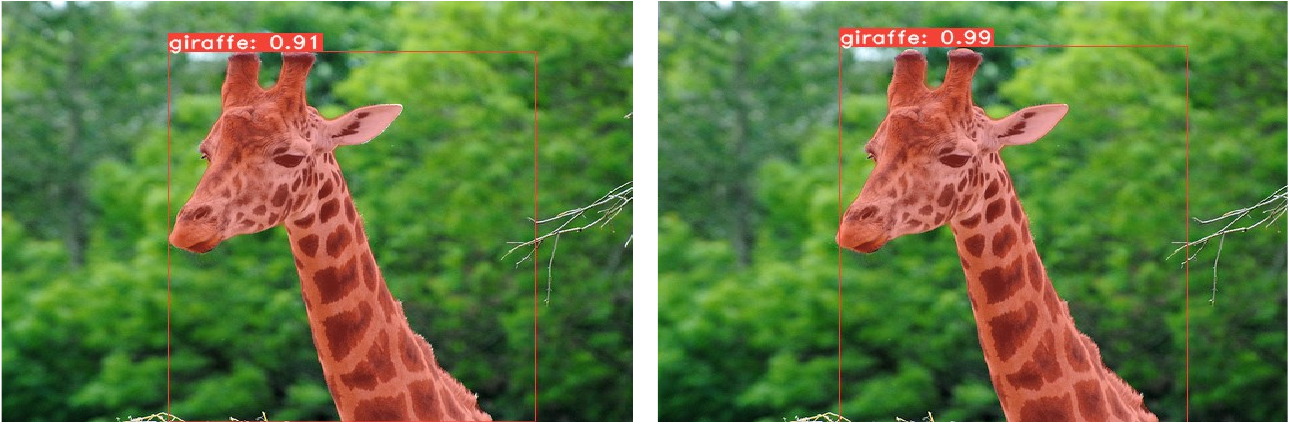}\label{fig:compare1}
     \end{minipage}}
     \hfill
     \subfigure[]{
     \begin{minipage}[t]{0.3\linewidth}
     \centering
    \includegraphics[scale=0.25]{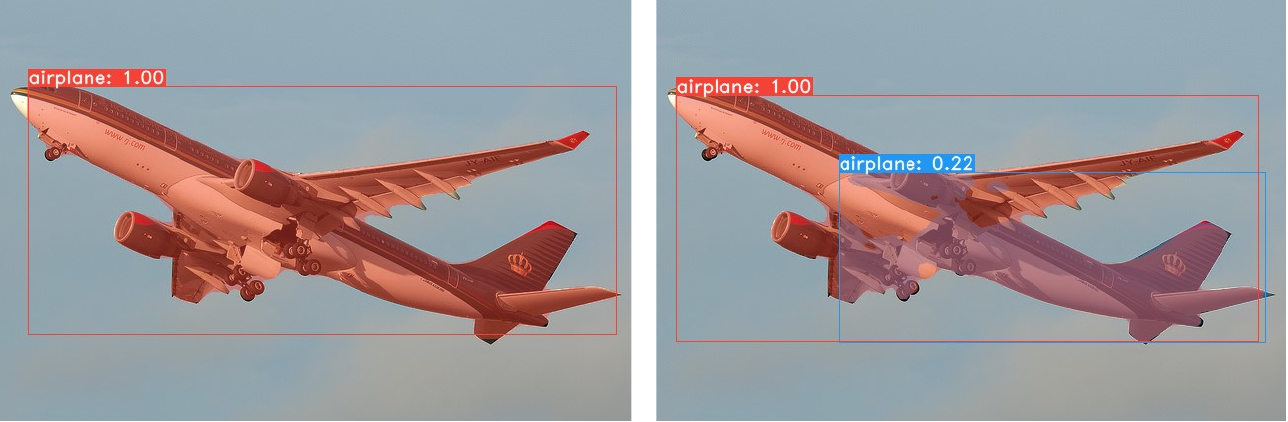}\label{fig:compare2}
     \end{minipage}}
      \hfill
      \subfigure[]{
      \begin{minipage}[t]{0.3\linewidth}
      \centering
      \includegraphics[scale=0.25]{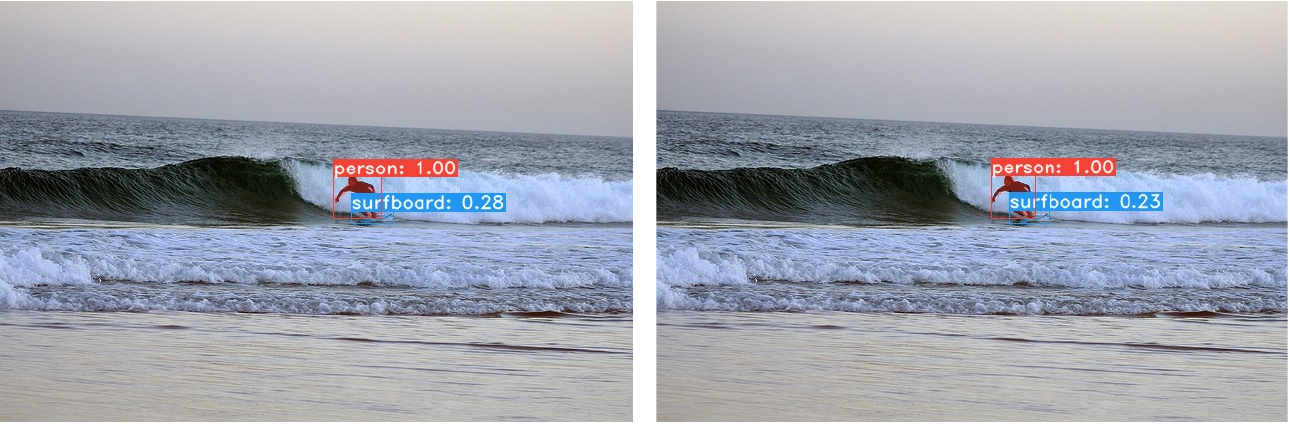}\label{fig:compare3}
      \end{minipage}}
     \caption{Detection examples using Faster R-CNN  trained on COCO 2017 dataset. Visualization samples are chosen from COCO test-2014. (a,b): Left: $\mathcal{L}_{\text{CIOU}}$, right: $\mathcal{L}_{\text{Focal-EIOU}}$. (c): Left: $\mathcal{L}_{\text{IOU}}$, right: $\mathcal{L}_{\text{Focal-EIOU}}$}
  \end{figure*}

 \begin{figure*}
  \centering
  \subfigure[]{
    \begin{minipage}[t]{0.32\linewidth}
    \centering
    \includegraphics[scale=0.3]{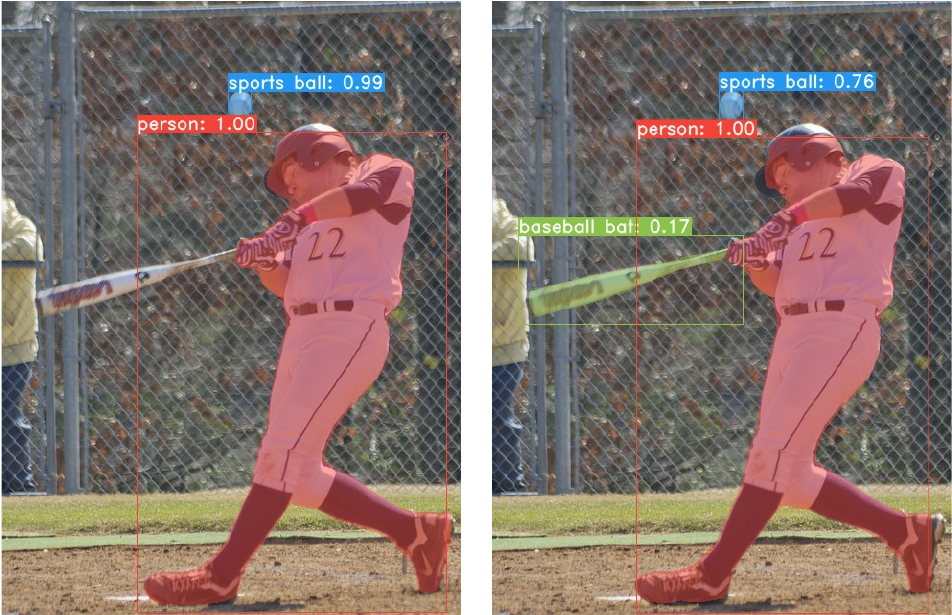}\label{fig:compare11}
     \end{minipage}}
     \hfill
     \subfigure[]{
     \begin{minipage}[t]{0.32\linewidth}
     \centering
    \includegraphics[scale=0.3]{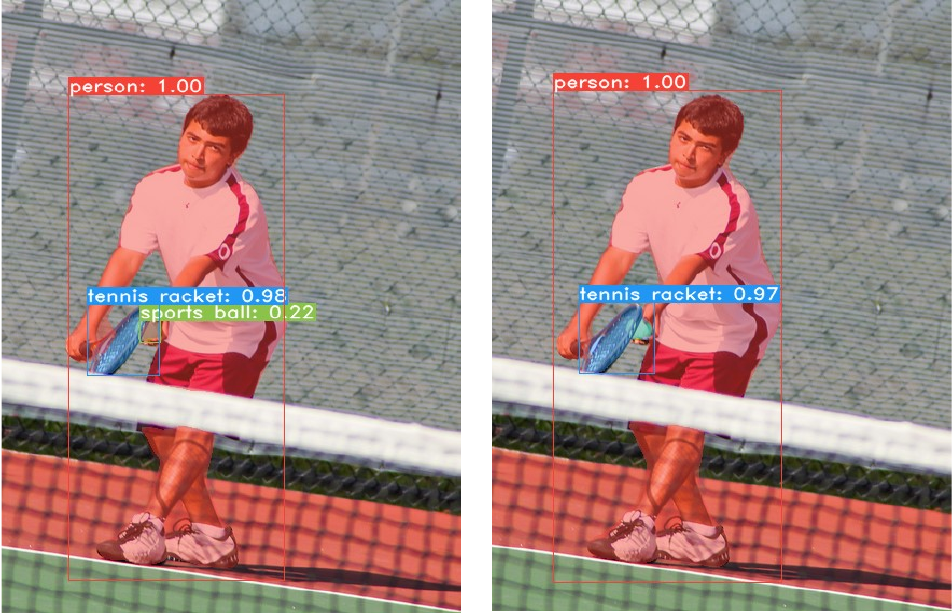}\label{fig:compare22}
     \end{minipage}}
      \hfill
      \subfigure[]{
      \begin{minipage}[t]{0.32\linewidth}
      \centering
      \includegraphics[scale=0.3]{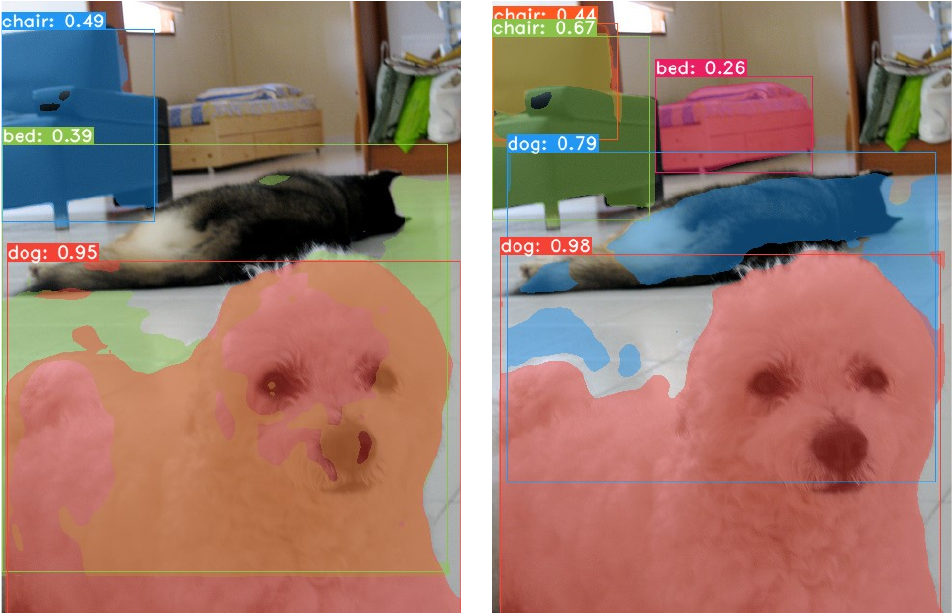}\label{fig:compare31}
      \end{minipage}}
      
     \caption{Detection examples using Faster R-CNN  trained on COCO 2017 dataset. Visualization samples are chosen from COCO test-2014. Left: $\mathcal{L}_{\text{CIOU}}$, right: $\mathcal{L}_{\text{Focal-EIOU}}$.}\label{fig:compare}
  \end{figure*}
\subsection{Discussion on Focal-EIOU loss and error set analysis}
\figurename~\ref{fig:compare1} shows that one can easily find more accurate boxes with higher prediction confidence. However, Focal-EIOU loss is more sensitive to large elements, it occasionally assigns wrong boxes near the large element, which is shown in~\figurename~\ref{fig:compare2}. Besides, in terms of the small elements, Focal-EIOU loss is a little inferior to the original IOU loss. In \figurename~\ref{fig:compare3}, Focal-EIOU loss assigns blurry boxes with lower prediction confidence to the small surfboard. Nevertheless, Focal-EIOU loss performs much better for medium and large objects. Samples in~\figurename~\ref{fig:compare} show that the proposed Focal-EIOU loss performs better on medium and large objects and may ignore or assign low-quality boxes and low-confidence predictions to small objects.

\section{Conclusion}
This paper takes a thorough analysis of BBR for object detection and find that the potential of the loss function has not been fully exploited. The first reason is that existing loss functions all have some drawbacks, hindering the regression of bounding boxes from being guided correctly. Secondly, exiting studies mostly neglect the importance of effective example mining of BBR. Therefore, low-quality examples contribute excessively large gradients and further limit the performance of BBR. Based on the observation, we propose the Focal-EIOU loss to tackle existing losses' defects and balance the gradients derived by the high and low-quality examples. Extensive experiments on both the synthetic and COCO dataset show that the Focal-EIOU loss brings significant and consistent improvements with a number of state-of-the-art models. 

{\small
\bibliographystyle{IEEEtrans}
\bibliography{egbib}
}
\end{document}